\documentclass[letterpaper, 10 pt, conference]{ieeeconf}  
\IEEEoverridecommandlockouts
\usepackage{cite}
\usepackage{amsmath,amssymb,amsfonts}
\usepackage{algorithm}
\usepackage{booktabs} 
\usepackage{algorithm}
\usepackage{algpseudocode}
\usepackage{graphicx}
\usepackage{textcomp}
\usepackage{xcolor}
\usepackage{subfiles}
\usepackage{mathtools}
\usepackage{subcaption}
\usepackage{graphicx}
\usepackage{svg}
\usepackage{url}
\usepackage{hyperref}
\newcommand\norm[1]{\lVert#1\rVert}

\def\BibTeX{{\rm B\kern-.05em{\sc i\kern-.025em b}\kern-.08em
    T\kern-.1667em\lower.7ex\hbox{E}\kern-.125emX}}

\begin{document}
\title{\LARGE \bf SAFE-TAXI: A Hierarchical Multi-UAS Safe Auto-Taxiing Framework with Runtime Safety Assurance and Conflict Resolution}
\author{Kartik A. Pant, Li-Yu Lin, Worawis Sribunma, Sabine Brunswicker, James M. Goppert and Inseok Hwang
\thanks{Kartik A. Pant, Li-Yu Lin, Worawis Sribunma, Inseok Hwang and James M. Goppert are with the School of Aeronautics and Astronautics, Purdue University,
West Lafayette, IN 47906. Email: \{{kpant}, {lin1191}, {wsribunm},{jgoppert},{ihwang}\}@purdue.edu}%
\thanks{Sabine Brunswicker is with the Polytechnic Institute and the Founding Director of $AIDA^3$ in Discovery Park, Purdue University,
West Lafayette, IN 47906. Email: sbrunswi@purdue.edu}
\thanks{A part of this research has been supported by a donation from Windracers LLC in the form of a Windracers Fellowship,
and the Purdue-Windracers Center $AIDA^3$, a Center for AI on Digital, Autonomous and Augmented Aviation
(\url{https://www.purdue.edu/computes/aida3/}).}
}

\maketitle
\begin{abstract}
We present a hierarchical safe auto-taxiing framework to enhance the automated ground operations of multiple unmanned aircraft systems (multi-UAS). The auto-taxiing problem becomes particularly challenging due to (i) unknown disturbances, such as crosswind affecting the aircraft dynamics, (ii) taxiway incursions due to unplanned obstacles, and (iii) spatiotemporal conflicts at the intersections between multiple entry points in the taxiway. To address these issues, we propose a \textit{hierarchical} framework, i.e., SAFE-TAXI, combining centralized spatiotemporal planning with decentralized MPC-CBF-based control to safely navigate the aircraft through the taxiway while avoiding intersection conflicts and unplanned obstacles (e.g., other aircraft or ground vehicles). Our proposed framework decouples the auto-taxiing problem temporally into conflict resolution and motion planning, respectively. Conflict resolution is handled in a centralized manner by computing conflict-aware reference trajectories for each aircraft. In contrast, safety assurance from unplanned obstacles is handled by an MPC-CBF-based controller implemented in a decentralized manner. We demonstrate the effectiveness of our proposed framework through numerical simulations and experimentally validate it using Night Vapor, a small-scale fixed-wing test platform.   
\end{abstract}

\section{Introduction}
\label{sec:intro}
Multi-UAS systems offer a low-cost, versatile, and energy-efficient solution for wide-ranging applications such as surveillance \cite{beard2006decentralized}, search and rescue \cite{erdos2013experimental}, and disaster management\cite{erdelj2017help}. For example, Windracers' mid-range autonomous aircraft ULTRA (Uncrewed Low-cost Transport) \cite{windracers2024ultra} has been built and used for parcel delivery and humanitarian support missions. The current major challenge in deploying these autonomous multi-UAS fleets for large-scale operations is their dependence on multiple human operators to accomplish their missions (up to three persons are required to taxi one aircraft). 
To scale the operation of multi-UAS and improve their efficiency, it becomes paramount to automate some of their regular ground operations, e.g., taxiing to/from the hanger to the runway. 

Multi-UAS auto-taxiing is an active area of research in both the control and aviation industry \cite{zhang2020software,lu2016improved,gaikwad2023developing}. There are several challenges associated with planning and coordinating multiple aircraft in the airport area, e.g., conflict-aware planning and scheduling, sensing and obstacle detection, localization, and collision avoidance, which require persistent human intervention. In this work, however, we focus on certain key aspects of multi-UAS taxiing. These are (i) automatic motion planning and control from the hanger to the runway and back, (ii) conflict resolution at the intersection of the entry point, and (iii) collision avoidance with unplanned obstacles (e.g., other aircraft or ground vehicles).
\begin{figure}[t]
\centering
\includegraphics[width=0.47\textwidth]{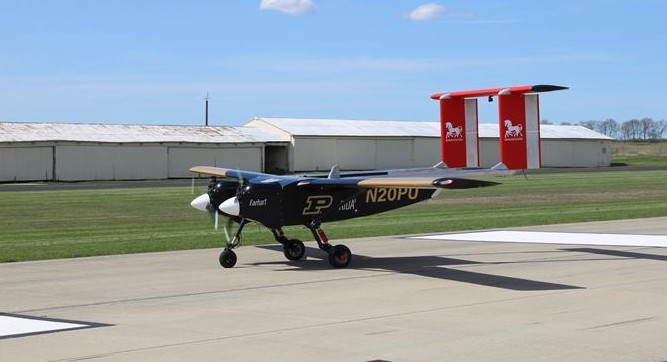}
\caption{Illustration of a mid-range fixed-wing aircraft (Windracers ULTRA) taxiing from the hanger to the runway.}
\label{fig:mixed_reality}
\end{figure}

\begin{figure*}[t]
\centering
\includegraphics[width=0.85\textwidth]{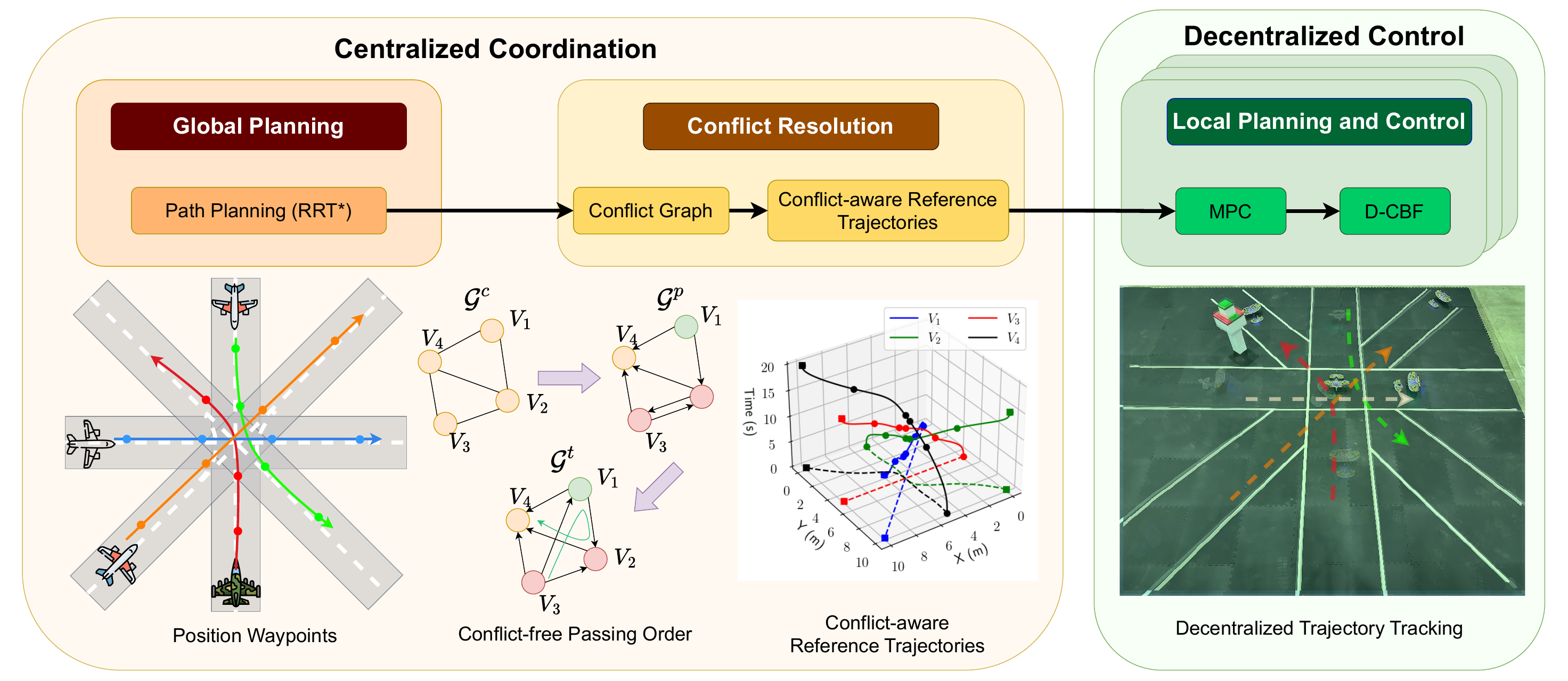}
\caption{Overview of our proposed SAFE-TAXI framework. Using the position waypoints from a global path planner, our framework first constructs a conflict graph among aircraft with spatial conflicts while crossing an intersection. The conflict graph is resolved to generate a passing order and desired time slots for each aircraft crossing the intersection. Conflict-aware reference trajectories are then computed by solving a polynomial trajectory optimization with desired time slots as boundary conditions. Finally, utilizing a decentralized MPC-CBF controller, each aircraft tracks the conflict-aware reference trajectories, thus ensuring the safety of the aircraft against unplanned taxiway incursions. }
\label{fig:arch}
\end{figure*}
To address these above-mentioned challenges, we present a hierarchical framework for runtime safety assurance and conflict resolution in multi-UAS auto-taxiing. Our goal is to decouple the overall problem temporally into conflict resolution and motion planning phases. In the centralized coordination phase, we design conflict-aware reference trajectories for each aircraft to allow it to navigate the taxiway without conflicts and negotiations with other aircraft. Meanwhile, in the motion planning phase, we design control inputs for each aircraft to track conflict-aware reference trajectories in a decentralized manner. We take advantage of the MPC-based receding horizon approach for control design. Furthermore, we incorporate dynamic control barrier functions (D-CBFs) as safety constraints to ensure the aircraft's runtime safety against unplanned taxiway incursions that may be caused by other aircraft, ground vehicles, humans, etc. Finally, we deploy our framework on an aircraft test platform, Night Vapor, to demonstrate its real-world application and feasibility. Furthermore, we integrate our entire software and hardware stack with a remote operations center, which we denote as the Smart Operation Center (SOC), for its validation by a human operator in real-time. 

In summary, the contributions of this paper are as follows:
\begin{itemize}
    \item We present a hierarchical multi-UAS auto-taxiing framework with runtime safety assurance and spatiotemporal conflict resolution to ensure the safe navigation of multiple aircraft in the taxiway.
    \item We experimentally demonstrate the effectiveness of our proposed approach using Night Vapor, an experimental fixed-wing test platform. 
    \item We augment our framework with a remote operation center, i.e., SOC, for its real-world applications. The SOC provides real-time command and control (C\&C), enabling human-in-the-loop validation of multi-UAS operations.
\end{itemize}
The remainder of the paper is organized as follows. Section \ref{sec:related_work} reviews related work on multi-UAS planning and runtime safety assurance of autonomous systems. Section \ref{sec:mpc_cbf} presents our proposed hierarchical framework focused on enhancing safety while maintaining the performance of multi-UAS auto-taxiing. Section \ref{sec:exp} presents the experimental validation of the framework. Finally, Section \ref{sec:conclusion} concludes the paper and outlines future research directions.

\section{Related Works}
\label{sec:related_work}
\subsection{Multi-UAS Taxi Planning and Routing}
A significant body of aviation and control literature has attempted to automate the routing and planning of multi-UAS taxiing. In \cite{zhao2021research}, an aircraft taxiing approach is proposed using a conflict-aware A*. A Monte Carlo Tree Search (MCTS) based framework is proposed in \cite{sui2023conflict} to address conflict resolution in auto-taxiing. In \cite{yang2021holistic}, the auto-taxiing problem is posed as a routing problem where \textit{Standardized Taxiing Routes} are established to reduce traffic congestion and taxiway incursions. The authors in \cite{deng2022multi} proposed particle swarm optimization-based methods for airport runway planning and conflict resolution. An approach based on graph neural networks (GNNs) is proposed in \cite{wang2024quick} to predict the optimal taxi schedule, and a model based on GNN-LSTM is used in \cite{yuan2025prediction} to resolve conflicts. 
Most of the above methods are suitable for manned aircraft settings, where the routing and planning are performed only once, and the human pilot ensures the aircraft's run-time safety. However, as we enter the realm of autonomous UAS, a unified framework that combines taxi planning and conflict resolution along with run-time safety assurance and collision avoidance becomes essential.


\subsection{Runtime Safety Assurance of Autonomous Systems}
Autonomous aircraft operations are safety-critical in nature, as any deviation from planned operations could lead to catastrophic results. 
To avoid such situations, several efficient methods have been proposed in the control theory and robotics literature that ensure the safety of the system. Among various methods, a popular class of approaches includes characterizing a set of system states that are deemed safe, i.e., safe sets, even in the worst case. These methods include control barrier function (CBF) \cite{ames2019control, zeng2021safety}, Hamilton-Jacobi-Bellman (HJB) reachability analysis \cite{bansal2017hamilton}, etc. Each method enforces the safety of the system but has its own advantages and disadvantages. CBF-based control design ensures the safety of the system by enforcing its state to stay in an invariant set characterized by a barrier function. On the other hand, the HJB method computes the system's exact Backward Reachable Set (BRS) from the unsafe sets by solving the Hamilton-Jacobi-Bellman partial differential equation to compute the safe sets. It provides a stricter characterization of the system's safety compared to the CBF approach; however, the main drawback of the HJB method is its high computational cost. The CBF-based approach is computationally superior as it can be easily integrated within a linear or quadratic program.
It relies on a nominal controller (designed to stabilize the system) that is minimally modified to satisfy the barrier function constraints, thereby ensuring set invariance of the system state to a safe set. 

Model predictive control (MPC) has recently gained popularity in the robotics community for its performance, state and input constraint handling, and robustness to disturbances and modeling errors. 
Recent efforts have unified the predictive capabilities of MPC with the safety certification of CBF \cite{zeng2021safety, zeng2021enhancing} to design controllers in a receding horizon fashion, enhancing system safety without sacrificing system performance. Several works in the robotics literature have utilized the MPC-CBF framework for various applications ranging from mobile robots\cite{jian2023dynamic, lu2024robot}, unmanned surface vehicles \cite{wang2024robust}, and unmanned aerial vehicles \cite{ali2024mpc,butler2024safe}. 

In \cite{butler2024safe}, the authors proposed an MPC-CBF approach for single aircraft taxiing. However, their method only focuses on static obstacle avoidance and does not consider inter-aircraft intersection conflicts, which can significantly disrupt multi-UAS operations. Moreover, the proposed method has not been implemented in a real-world scenario, thus undermining its applicability. To address these challenges, in this paper, we consider a unified framework for multi-UAS auto-taxiing, combining collision avoidance and inter-aircraft conflict resolution simultaneously to minimize intersection conflicts and avoid unplanned taxiway incursions, while ensuring the aircraft's overall performance, safety, and fuel efficiency.

\section{Heirarchical Safe Autotaxiing Framework}
\label{sec:mpc_cbf}
\begin{figure*}[t]
\centering
\includegraphics[width=\textwidth]{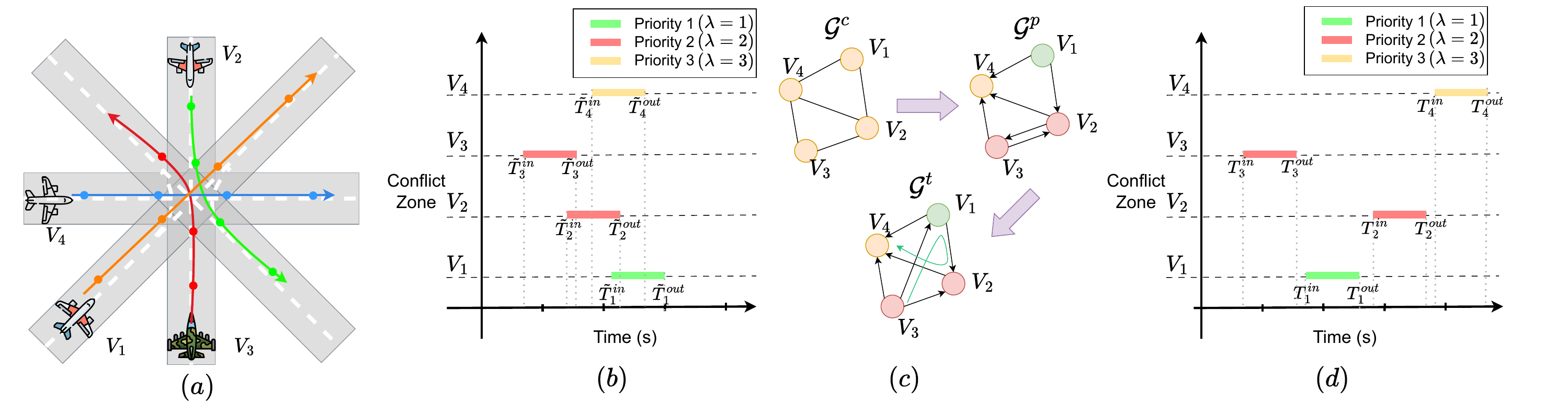}
\caption{(a) An example scenario of multi-UAS auto-taxiing showing $4$ aircraft approaching an 8-way intersection with each aircraft in spatial conflict with one another, (b) Initial time slots estimated using the global path planner's position waypoints and the operating speed of the aircraft, (c) Construction and sequential resolution of a conflict graph to obtain conflict-free passing order, and (d) Final conflict-free time slots for each aircraft after conflict resolution.}
\label{fig:conflict_res}
\end{figure*}
In this section, we first describe the aircraft dynamics and obstacles in Section \ref{subsec:model} and then present the overall hierarchical framework for multi-UAS auto-taxiing. Our proposed framework assumes that a global path planner, such as RRT* \cite{karaman2011sampling}, provides position waypoints for each vehicle. In Section \ref{subsec:conflict}, we explain the method of obtaining a conflict-free passing order and the desired time slots of all aircraft passing through an intersection. We optimize a polynomial trajectory for each aircraft using the conflict-free time slots, which is described in Section \ref{subsec:reference}. Finally, we explain the details of the decentralized MPC-CBF controller in Section \ref{subsec:local} that tracks conflict-aware reference trajectories and ensures aircraft safety against unplanned taxiway incursions. The overall architecture of the system is shown in Fig. \ref{fig:arch}
\subsection{Aircraft and Obstacle Model}
\label{subsec:model}
Suppose there are $n$ aircraft and $m$ obstacles on the taxiway. We model the motion of each aircraft $i \in \{1, \dots, n\}$ as a \textit{non-holonomic car}, as follows:
\begin{equation}
\label{eq:sys_model}
    \mathbf{x}_{i,k+1} = f_i(\mathbf{x}_{i,k}, \mathbf{u}_{i,k}) \coloneqq \begin{bmatrix}
        p^x_{i,k} \\
        p^y_{i,k} \\
        \theta_{i,k} \\
        v_{i,k} 
    \end{bmatrix} + \begin{bmatrix}
        v_{i,k} \cos \theta_k \\
        v_{i,k} \sin \theta_k \\
        \frac{v_{i,k}}{L} \tan \phi_{i,k} \\
        \beta_{i,k}
     \end{bmatrix} \Delta T,
\end{equation}
where $\mathbf{x}_{i,k} \coloneqq \begin{bmatrix} p^x_{i,k} & p^y_{i,k} &\theta_{i,k} & v_{i,k} \end{bmatrix}^\top$ is the state, $\mathbf{u}_{i,k} \coloneqq \begin{bmatrix} \phi_{i,k} & \beta_{i,k} \end{bmatrix}^\top$ is the control input of aircraft $i$ at time $k$, respectively. $L_i$ denotes the length, $ p^x_{i,k}$ and $ \ p^y_{i,k}$ denote the $x-y$ position, $v_{i,k}$ denotes the forward velocity, $\theta_{i,k}$ denotes the heading, $\phi_{i,k}$ denotes the rudder deflection angle, and $\beta_{i,k}$ denotes the throttle of vehicle $i$, respectively. All aircraft are subject to state and input constraints, i.e., $(\mathbf{x}_{i,k},\mathbf{u}_{i,k})  \in \mathcal{Z}_i$ for all $k\geq 0$, where $\mathcal{Z}_i \subset \mathbb{R}^{6}$ is a compact set. The aircraft model in \eqref{eq:sys_model} is adopted from various studies in the literature focused on aircraft auto-taxiing \cite{liu2019vision}.

Consider the evolution of the position of obstacle (including other aircraft) $j \in \{1, \dots, m\}$ as follows:
\begin{equation}
\label{eq:obs_model}
    \mathbf{o}_{j,k+1} = \xi(\mathbf{o}_{j,k}), 
\end{equation}
where $\mathbf{o}_{j,k+1} \in \mathbb{R}^2$ denotes the position of obstacle $j$ at time $k$ and $\xi(\cdot)$ is a Lipschitz continuous nonlinear function. For simplicity, we assume that the shape of each obstacle can be approximated as a circle of radius $r_j$ centered at $\mathbf{o}_{j,k}$. 
\subsection{Conflict Resolution via Spatiotemporal Planning}
\label{subsec:conflict}
Using position waypoints and the initial time slots of each aircraft approaching an intersection, we first design a passing order for all aircraft. 
Our design philosophy is to ensure that the \emph{aircraft that arrives first must leave first} unless one has a higher priority than the other. 

We define the following notation required to describe our approach. Let $n_i$ be the number of aircraft approaching an intersection, and let $\lambda_i$ denote the priority of each aircraft. We denote $[\tilde{T}_{i}^{in} \ \tilde{T}_{i}^{out}]$ as the initial intersection time slots computed using the position waypoints and the operating speed of the aircraft and $[T_{i}^{in} \ T_{i}^{out}]$ as the time slots after the resolution of the conflict at the intersection. We aim to find the time slots for each vehicle that avoid conflicts and then use them to construct conflict-aware reference trajectories that minimize snap/jerk for each aircraft. This ensures that aircraft do not stop abruptly in the conflict zone nor navigate in the taxiway with large variations in control input. These trajectories are subsequently tracked by an MPC-CBF-based decentralized controller, which is described later. 

For each conflict zone, we first construct a conflict graph by adding links between the pair of aircraft with spatial conflicts. We denote $\mathcal{U}_i$ as the set of aircraft that are in spatial conflict with aircraft $i$ as:
\begin{equation}
    \mathcal{U}_i \coloneqq \{j\ | \ \tilde{T}_{i}^{in} < \tilde{T}_{j}^{in} \leq \tilde{T}_{i}^{out}\},
\end{equation}
where $[\tilde{T}_{i}^{in}\ \tilde{T}_{i}^{out}]$ denotes the time slots estimated based on the estimated time-of-arrival (ETA) computed using global planner's position waypoints and operating speeds of the aircraft. The undirected graph constructed using the link between $i$ and $j$, defined as $\mathcal{G}^{c}\coloneqq \bigcup_i \bigcup_{j\in \mathcal{U}_i}(i,j)$, is a conflict graph \cite{guberinic2007optimal}. Finding the optimal passing order for a general conflict graph is an NP-Hard problem \cite{liu2017distributed}. In this paper, however, we consider a heuristic conflict resolution approach to avoid intersection conflicts by leveraging the unique structural properties of multi-UAS auto-taxiing problem. Since auto-taxiing is performed in an airport-like environment, aspects such as taxiway intersections, operating speeds, and priorities of each vehicle are already characterized and regulated. We take advantage of this to resolve the conflict graph in a computationally efficient manner as follows.
\begin{algorithm}[h]
\caption{Passing Sequence Using Conflict Graph}
\label{alg:passing_order}
\begin{algorithmic}[1]
\Require Conflict graph $\mathcal{G}^c$, Priority order $\Lambda = \{\lambda_1, \dots, \lambda_{n_i}\}$
\State Construct graph $\mathcal{G}^p$ by converting $\mathcal{G}^c$ to a directed graph using $\Lambda$.
\State Resolve double edges in $\mathcal{G}^p$ by prioritizing the aircraft with earlier arrival times to construct a temporal advantage graph $\mathcal{G}^t$.
\State Create additional directed edges between disjoint edges based on their updated time slots.
\State Find the longest walk in the temporal advantage graph $\mathcal{G}^t$ from the first aircraft arriving at the intersection.
\State The sequence of vertices in the longest walk is the resulting conflict-free passing order $\mathbf{v}^*$.
\State \textbf{Return} \( \mathbf{v}^* \)
\end{algorithmic}
\end{algorithm}

Given a conflict graph $\mathcal{G}^c$, we construct two additional graphs as shown in Fig. \ref{fig:conflict_res}c, a priority graph $\mathcal{G}^p$ and a temporal advantage graph $\mathcal{G}^t$, respectively. The graph $\mathcal{G}^p$ is a directed version of graph $\mathcal{G}^c$, where each directed edge $(i,j)$ represents the priority of aircraft $i$ over $j$ by comparing $\lambda_i$ and $\lambda_j$. If both aircraft have the same priority, we create a double edge between them. The temporal advantage graph $\mathcal{G}^t$ resolves double edges by prioritizing the aircraft whose arrival time to the intersection is earlier than the other. Based on the earliest arrival time principle, it also creates a directed edge between two disjoint vertices representing aircraft that do not have any spatial conflict with each other. In the case of equality, the tie-breaking is performed randomly. Using the temporal advantage graph, we can obtain the passing order by traversing the graph and obtain the largest walk starting from the aircraft that first arrives at the intersection, as shown in the bottom of Fig. \ref{fig:conflict_res}c. We define the passing order as $\mathbf{v}^* \in \mathbb{R}^{n_i}$. The algorithm for computing $\mathbf{v}^*$ is presented in Alg. \ref{alg:passing_order}. Note that the above algorithm is designed to resolve only local conflicts at each intersection. We posit that global conflict resolution, considering multiple intersections simultaneously, is a considerably challenging problem which we wish to defer as future work.    

Now, we design the timeslots for each aircraft using the locally optimal passing sequence. We denote $\Delta T_{safe}$ as the safe time added to enhance the robustness of our proposed approach. The time slot for each aircraft can be computed as: 
\begin{equation}
\label{eq:timeslots}
    T_{\mathbf{v}^*[i]}^{in}  = \begin{cases}
      \tilde{T}_{\mathbf{v}^*[0]}^{in} & \text{if $i=0$}\\
      T_{\mathbf{v}^*[i-1]}^{in} + \Delta T_{safe} &  \text{otherwise}
    \end{cases}  
\end{equation}
where $\mathbf{v}^*$ is the passing order for the aircraft and $T_{\mathbf{v}^*[i]}^{in}$ is the entering time of aircraft $\mathbf{v}^*[i]$ into a conflict zone. The time of the first aircraft approaching the intersection is considered a reference, and a schedule is formed for the rest of the aircraft using \eqref{eq:timeslots}, as shown in Fig. \ref{fig:conflict_res}d.
\subsection{Conflict-aware Reference Trajectories}
\label{subsec:reference}
Using position waypoints and updated time slots from the heuristic presented earlier, we construct the reference trajectory for each aircraft by leveraging polynomial trajectory optimization \cite{richter2016polynomial}. Given the position waypoints and their times, we obtain a minimum snap trajectory by solving the following optimization problem:
\begin{equation}
\label{eq:traj_opti}
    J_{r}^* = \min_{\mathbf{a}} \int_{t_0}^{t_f} \big(P^{r}(t)\big)^2 dt
\end{equation}
where $J_{r}^*$ denotes the optimal cost, $P: [t_0, t_f]\rightarrow \mathbb{R}$ denotes a continuously and sufficiently differential curve. $r$ is the derivative order that is minimized, e.g., $r=4$ implies that the snap is minimized. $\mathbf{a} = [a_1, \dots, a_{N_p}]$ denotes the coefficients of the curve $P$ characterized by the basis set $[1, t, t^2, \dots, t^{N_p-1}]$, where $N_p$ is the order of the polynomial. The position waypoints and their new time slots after the conflict resolution step are considered as the boundary conditions. The detailed methodology for the computation of the reference trajectories is beyond the scope of this work (see \cite{richter2016polynomial} for more details).
\begin{figure}[t]
\centering
\includegraphics[width=0.3\textwidth, trim ={0.5cm 0.5cm 0.5cm 0.5cm}, clip]{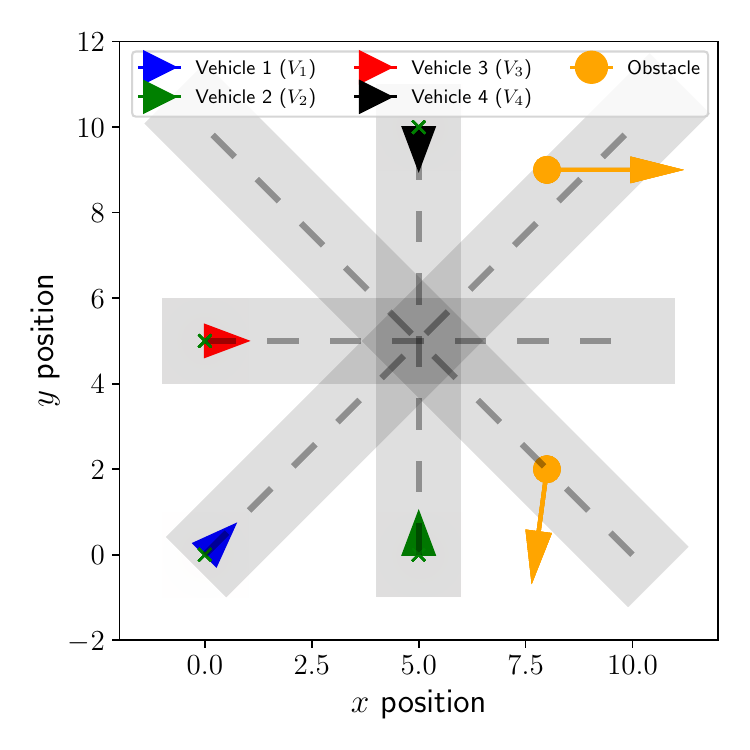}
\caption{Simulation scenario for validation of our proposed SAFE-TAXI framework.}
\label{fig:scenario}
\end{figure}
The resulting reference trajectories are denoted as $\bar{x}_i$ for each aircraft $i$. These reference trajectories ensure that all aircraft cross the intersection based on their respective time slots, minimizing the risk of conflict and negotiation.
\subsection{Local Planning and Control}
\label{subsec:local}
We use the MPC-CBF framework for local planning and control. Our main goal is to design a decentralized tracking controller to track the conflict-aware reference trajectories $\bar{x}_i$ generated by \eqref{eq:traj_opti}, which simultaneously avoid collisions with unplanned obstacles and other vehicles. 
Our approach is based on the recently proposed dynamic-CBF (D-CBF) \cite{jian2023dynamic, lu2024robot} for dynamic collision avoidance in mobile robots. 
 
The main idea is to construct safe sets $\mathcal{C}_{i,k}^j \subset \mathbb{R}^4 \times \mathbb{R}^{2}$ for each aircraft $i$ corresponding to each obstacle $j$, i.e., a 0-superlevel set of a continuously differential function $h_i^j: \mathbb{R}^4 \times \mathbb{R}^{2} \rightarrow \mathbb{R}$. It is defined as follows:
\begin{align}
\label{eq:safe_set}
    \mathcal{C}_{i,k}^j &= \{(\mathbf{x}_{i,k},\mathbf{o}_{j,k}) \subseteq \mathbb{R}^4 \times \mathbb{R}^{2} \ :  h_i^j(\mathbf{x}_{i,k},\mathbf{o}_{j,k})\geq 0 \}\\
    \partial\mathcal{C}_{i,k}^j &= \{\mathbf{x}_{i,k},\mathbf{o}_{j,k}) \subseteq \mathbb{R}^4 \times \mathbb{R}^{2} \ :  h_i^j(\mathbf{x}_{i,k},\mathbf{o}_{j,k})= 0 \}
\end{align}

Then, according to the definition of a discrete-time CBF \cite{agrawal2017discrete}, the safe set $\mathcal{C}_{i,k}^j$ is a forward invariant set, if and only if there exists a class $\mathcal{K}_\infty$ function $\alpha(\cdot)$ such that 
\begin{equation}
    \inf_{\mathbf{u}_i}\Delta  h_i^j(\mathbf{x}_{i,k},\mathbf{o}_{j,k},\mathbf{u}_{i,k}) \geq -\alpha( h_i^j(\mathbf{x}_{i,k},\mathbf{o}_{j,k})),
\end{equation}
where $\Delta  h_i^j(\mathbf{x}_{i,k},\mathbf{o}_{j,k},\mathbf{u}_{i,k})\coloneqq h_i^j(\mathbf{x}_{i,k+1},\mathbf{o}_{j,k+1}) -  h_i^j(\mathbf{x}_{i,k},\mathbf{o}_{j,k})$. Using the results from \cite{agrawal2017discrete}, we select the $\alpha( h_i^j(\mathbf{x}_{i,k},\mathbf{o}_{j,k}))$ as $\gamma h_i^j(\mathbf{x}_{i,k},\mathbf{o}_{j,k})$, where $0<\gamma\leq1$. Thus, the CBF constraint for each obstacle is defined as:
\begin{equation}
    \bar{h}_{i, k}^j \coloneqq h_i^j(\mathbf{x}_{i,k+1},\mathbf{o}_{j,k+1}) -  (1-\gamma) h_i^j(\mathbf{x}_{i,k},\mathbf{o}_{j,k}) \geq 0,
\end{equation}
where $\bar{h}_{i, k}^j = \bar{h}_{i, k}^j(\mathbf{x}_{i,k},\mathbf{o}_{j,k})$ denotes the D-CBF constraint.

Since we model each obstacle as a circle of radius $r_j$, we can formulate a discrete-time D-CBF for multi-UAS auto-taxiing in a quadratic form as:
\begin{equation}
    h_i^j(\mathbf{x}_{i,k},\mathbf{o}_{j,k}) = \norm{D_i \mathbf{x}_{i,k} - \mathbf{o}_{j,k}}^2 - (r_j + d_{safe})^2,
\end{equation}
where $D_i$ is a matrix which picks the position of the aircraft from its state $\mathbf{x}_{i,k}$ and $d_{safe}$ is the safety distance. We denote $\mathcal{C}_{i,k} = \bigcup_{\forall j \in{1,\dots, m}} \mathcal{C}_{i,k}^j$ as the union of the safe sets for each aircraft $i$.

We will now show how these constraints can be effectively introduced in the MPC optimization problem. Consider $\mathbf{x}_{i,l|k}$ and $\mathbf{u}_{i,l|k}$ as the $l^{\text{th}}$ step forward prediction of the state and inputs of aircraft $i$ at time $k$, respectively, with $l \in \{1,\dots, N\}$, where $\mathbf{x}_{i,0|k} = \mathbf{x}_{i,k}$ and $\mathbf{u}_i = [u_{i,0|k}^\top, u_{i,1|k}^\top, \dots, u_{i,N|k}^\top]^\top$. Let $N$ be the prediction horizon for the MPC. Then, the MPC optimization problem can be stated as:
\begin{subequations}
\label{eq:mpc}
\begin{align}
    \label{eq:mpc_cost}
    J_i^*(\mathbf{x}_{i,k}) = \min_{\mathbf{u}_{i}} \ &p(\mathbf{x}_{k+N|k}) + \sum_{l=0}^{N-1} q(\mathbf{x}_{i,k}, \mathbf{u}_{i,k})\\
    &\text{s.t.} \quad l = \{1,\dots, N-1\}\\
    \label{eq:mpc_init}
    &\mathbf{x}_{i,0|k} = \mathbf{x}_{i,k}\\
    \label{eq:mpc_dyn}
    &\mathbf{x}_{i,l+1|k} = f_i(\mathbf{x}_{i,l|k}, \mathbf{u}_{i,l|k})\\
    \label{eq:mpc_const}
    &(\mathbf{x}_{i,l|k}, \mathbf{u}_{i,l|k}) \in \mathcal{Z}_i\\
    \label{eq:mpc_inv}
    &\mathbf{x}_{i,N|k} \in \Gamma_i\\
    \label{eq:mpc_cbf}
    &\bar{h}_{i, k+l|k}^j \geq 0 \quad \forall j = \{1,\dots, m\}
\end{align}
\end{subequations}
where \eqref{eq:mpc_cost} is the cost function comprising a stage cost $\sum_{l=0}^{N-1} q(\mathbf{x}_{i,k}, \mathbf{u}_{i,k})$ and a terminal cost $p(\mathbf{x}_{k+N|k})$, \eqref{eq:mpc_init} is the initial condition, and \eqref{eq:mpc_const} are the state and input constraints, respectively. The terminal state constraint is denoted by \eqref{eq:mpc_inv}, where $\Gamma_i$ is the terminal invariant set. \eqref{eq:mpc_cbf} denotes the safety constraints for each obstacle that yields $\mathcal{C}_{i,k}$ as an invariant set. In this work, we consider the stage and terminal cost to have the following quadratic form defined as $p(\mathbf{x}_{k+N|k})\coloneqq \norm{\mathbf{x}_{i,N|k} - \bar{\mathbf{x}}_{i,N|k}}_{P}^2$ and the stage cost $q(\mathbf{x}_{i,k}, \mathbf{u}_{i,k})\coloneqq \norm{\mathbf{x}_{i,k} - \bar{\mathbf{x}}_{i,k}}_{Q}^2 +\norm{\mathbf{u}_{i,k+1} - \mathbf{u}_{i,k}}_{R}^2$, where $P,Q,R$ are positive-definite weighting matrices of appropriate dimensions.

\section{Experiments}
\label{sec:exp}
In this section, we validate the effectiveness of our proposed framework via both numerical simulations and hardware experiments using a fixed-wing experimental test platform, i.e., Night Vapor. Furthermore, we present the additional interface developed for the integration of our framework with a remote operation center, i.e., SOC, bolstering its utility for human-in-the-loop verification for future works. 

\textit{Scenario:} We consider a multi-UAS auto-taxiing scenario where $4$ aircraft cross an intersection, as shown in Fig. \ref{fig:scenario}. The aircraft $1$ and $4$ encounter unplanned obstacles on their way (shown in orange). The obstacles move in the direction of the arrow at a constant speed of $0.1$ m/s (chosen arbitrarily to make the scenario interesting). We assume that the dynamics of obstacle positions are known to the aircraft.  
\subsection{Numerical Simulations}
We implement our algorithm using CasADi \cite{andersson2019casadi}. The sample time is $\Delta T = 0.1s$. The acceleration and deflection of the rudder for each aircraft are limited to $\norm{\beta_i} \leq 1$ $m/s^2$ and $\norm{\phi_i} \leq \pi/6$ rad, respectively. For the D-CBF constraints, we chose $\gamma=0.1$, and the weights of the MPC cost functions are chosen as $Q = \text{diag}[10,10,5,5]$ and $R = \text{diag}[0.01, 0.1]$, respectively, and the MPC horizon length is chosen as $N =15$. We choose $N_p=8$ as the order for the reference trajectory polynomial.

To evaluate the performance of our proposed framework for the scenario described in the previous section, we compare it with (i) naive MPC-CBF (without conflict resolution) and (ii) MPC-CBF (with wait-and-go), in which each aircraft first stops at the intersection, breaks ties based on the aircraft's priority and then move. We adopt the following indicator to compare the performance of our proposed framework: (i) \emph{comp-time} (time spent from start to goal for all aircraft) and (ii) \emph{avg-acc-var} (average acceleration variance while avoiding obstacles and conflicts across number of vehicles). We use the same parameters defined earlier for all three approaches. The results are shown in Table \ref{tab:comparison}. 
\begin{table}[h]
\centering
    \begin{tabular}{cccc} 
        \hline
        & \textbf{Algorithms} & \textbf{Comp-time ($s$)} & \textbf{Avg-acc-var ($m/s^2$)} \\
        \hline
        & MPC-CBF (naive) & deadlock & - \\ 
        & MPC-CBF (wait-and-go) & 20.4 & 0.182 \\ 
        & SAFE-TAXI (ours) & \textbf{19.2} & \textbf{0.113}\\ 
        \hline
    \end{tabular} 
    \caption{Comparison with the baseline approaches}
    \label{tab:comparison}
\end{table}

The results clearly show that using proactive spatiotemporal planning, our proposed method performs better over the baseline, which is a reactive stop-wait-go approach to crossing an intersection (similar to an unmanaged intersection crossing in urban traffic scenarios) in both metrics. Moreover, our proposed approach has a $37\%$ lower acceleration variance compared to the baseline, which is critical for the lifecycle of an aircraft's brakes and other hardware components. Note that we did not consider friction in our simulations; however, in reality, static friction can yield significantly higher acceleration variance with the stop-wait-go approach.    
\subsection{Real-world Demonstration}
We perform real-world experiments using the Night Vapor platform to showcase the practical applications of our proposed SAFE-TAXI framework. We added a tail wheel (not a part of the off-the-shelf kit) to enhance the controllability of the Night Vapor while taxiing. Additionally, we added an Arduino-based pulse-position modulation (PPM) signal converter that converts MPC-CBF control commands to PWM signals. We conduct our experiments at the Purdue UAS Research and Test Facility (PURT). It houses the world's largest indoor motion capture with 60 Oqus 7+ Qualisys cameras distributed across 20,000 square feet of area. We use $4$ Night Vapor aircraft to demonstrate our proposed auto-taxiing framework. We demonstrate real-world experiments using the same scenario as discussed in the previous section. Each Night Vapor is controlled using an independent Linux computer which runs its own ROS2 instance, leveraging the ground-truth position measurements from the motion capture system. This showcases the decentralized aspect of our proposed framework. The control commands from the ROS2 node, i.e., acceleration and rudder deflection, are mapped to actuate the propeller and the rudder surface, respectively. Figure \ref{fig:real_exp}a shows that the four aircraft safely cross the intersection while tracking their individual conflict-aware reference trajectories. This validates the real-world implementation of our proposed hierarchical safe multi-UAS auto-taxiing framework. 
\begin{figure}[t]
\centering
\includegraphics[width=0.47\textwidth]{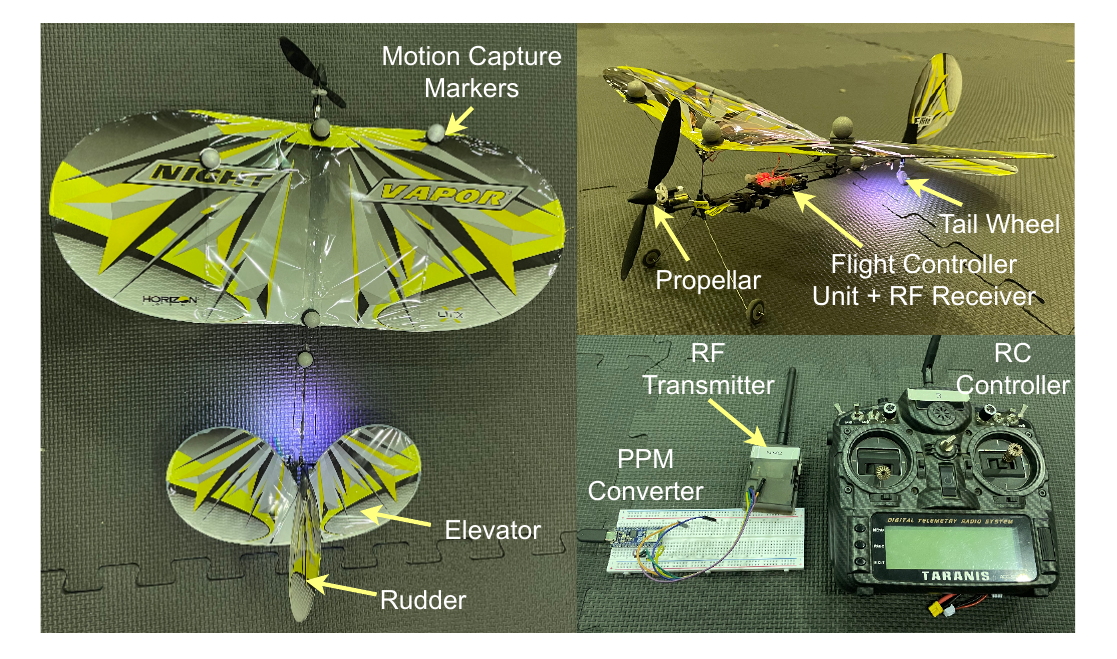}
\caption{Test platform for the experiment, Night Vapor, is a small-scale fixed-wing vehicle equipped with a flight controller unit, RF receiver and bind-and-play interface.}
\label{fig:nightvapor}
\end{figure}
\subsection{SOC Integration} 
\begin{figure*}[t]
\centering
\includegraphics[width=0.93\textwidth]{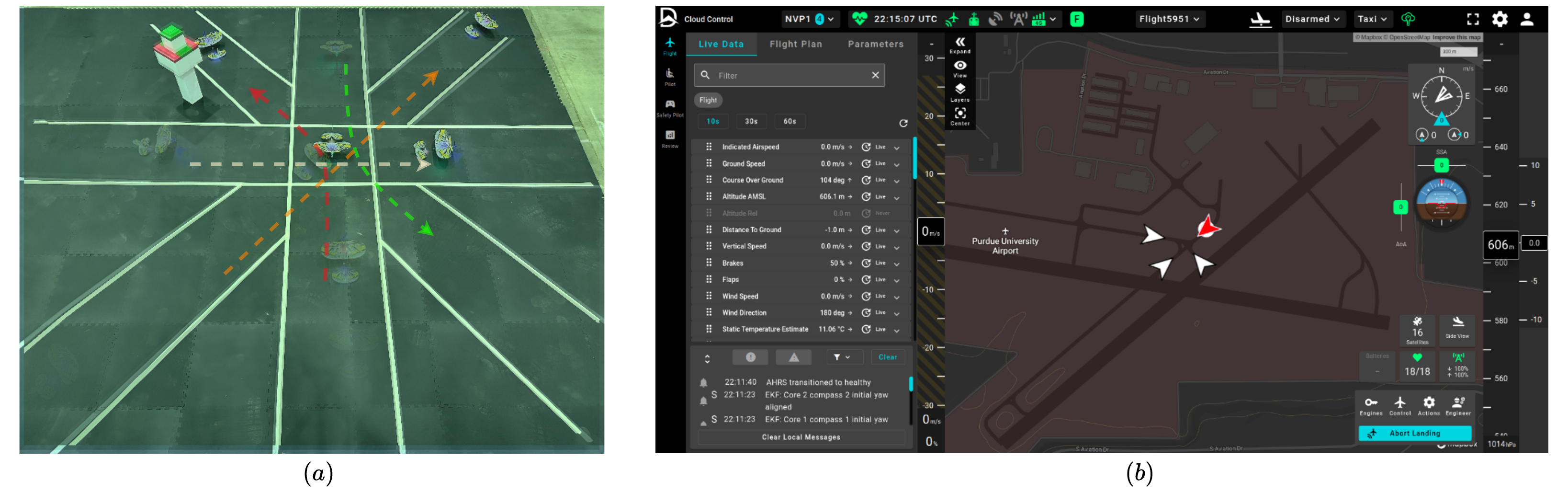}
\caption{(a) Real Experiments with $4$ Night Vapor vehicles executing auto-taxiing using our proposed SAFE-TAXI framework, (b) Graphical user interface console depicting the live locations (scaled) of all the aircraft performing auto-taxiing on a cloud-based real-time remote flight monitoring system.}
\label{fig:real_exp}
\end{figure*}

To further demonstrate the practical applications of our proposed framework, we integrate it with our existing testbed \cite{sribunma2024testbed} and connect it to a remote operation center, i.e., SOC, designed to operate large fixed-wing UAS. We develop additional interfaces using Pymavlink, a Python MAVLink \cite{meier2013mavlink} interface for UAS applications. We send scaled positions of our small-scale aircraft to the SOC. This enables real-time remote communication and authentication from human operators to execute/validate the autonomous operations of the multi-UAS auto-taxiing operations. Figure \ref{fig:real_exp}b shows the user interface of a custom cloud-based architecture showing the real-time locations of the UAS performing auto-taxiing. 

\section{Conclusion}
\label{sec:conclusion}
In this paper, we proposed a hierarchical framework for multi-UAS auto-taxiing, which ensures aircraft safety from unplanned taxiway incursions and allows them to navigate through the taxiway without any conflicts or negotiations. We showed that by leveraging the structural properties of the multi-UAS auto-taxiing problem, we can efficiently resolve the spatial conflicts among aircraft by using a conflict graph to generate conflict-aware reference trajectories. 
We validated the effectiveness of the proposed approach via both numerical simulations and experimental validation using a fixed-wing test platform, Night Vapor, and we demonstrated that our approach has lower acceleration variance than the baseline approaches. Finally, we integrated our framework with a remote operation center (SOC) to enable its human-in-the-loop validation capabilities. In the future, we would like to extend our work for handling multiple intersections simultaneously, thus further enhancing the performance of the auto-taxiing operation.

\bibliographystyle{ieeetr}
\bibliography{IEEEabrv,bibs} 

\end{document}